\documentclass[journal]{IEEEtran}
\usepackage{multirow} 
\usepackage{booktabs}
\usepackage{makecell}
\usepackage{cite}
\usepackage{amsmath} 
\usepackage[ruled]{algorithm2e} 
\usepackage{array}
\usepackage{graphicx}
\usepackage{float}
\usepackage{amsthm,amsmath,amssymb}
\usepackage[table,xcdraw]{xcolor}
\graphicspath{{./figures/}}
\usepackage{url}  

\ifCLASSOPTIONcompsoc
  \usepackage[caption=false,font=normalsize,labelfont=sf,textfont=sf]{subfig}
\else
  \usepackage[caption=false,font=footnotesize]{subfig}
\fi

\begin{document}
\title{UPNet: Uncertainty-based Picking Deep Learning Network for Robust First Break Picking}

\author{Hongtao~Wang, Jiangshe~Zhang, Xiaoli~Wei, Li~Long, Chunxia~Zhang
\thanks{Corresponding authors: Jiangshe Zhang}
\thanks{H.T. Wang, J.S. Zhang, X.L. Wei, L. Long, C.X. Zhang are with the School of Mathematics and Statistics, Xi'an Jiaotong University, Xi'an, Shaanxi, 710049, P.R.China.}
\thanks{The research is supported by the National Key Research and Development Program of China under grant 2020AAA0105601, the National Natural Science Foundation of China under grant 61976174 and 61877049.}}

\markboth{Journal of \LaTeX\ Class Files,~Vol.~14, No.~8, May~2023}%
{Wang \MakeLowercase{\textit{et al.}}: UPNet for Robust First Break Picking}

\maketitle

\begin{abstract}
In seismic exploration, first break (FB) picking is a crucial aspect in the determination of subsurface velocity models, significantly influencing the placement of wells. Many deep neural networks (DNNs)-based automatic picking methods have been proposed to accelerate this processing. Significantly, the segmentation-based DNN methods provide a segmentation map and then estimate FB from the map using a picking threshold. However, the uncertainty of the results picked by DNNs still needs to be analyzed. Thus, the automatic picking methods applied in field datasets can not ensure robustness, especially in the case of a low signal-to-noise ratio (SNR). In this paper, we introduce uncertainty quantification into the FB picking task and propose a novel uncertainty-based picking deep learning network called UPNet. UPNet not only estimates the uncertainty of network output but also can filter the pickings with low confidence. Many experiments evaluate that UPNet exhibits higher accuracy and robustness than the deterministic DNN-based model, achieving State-of-the-Art (SOTA) performance in field surveys. In addition, we verify that the measurement uncertainty is meaningful, which can provide a reference for human decision-making.

\end{abstract}

\IEEEpeerreviewmaketitle

\section{Introduction}
\IEEEPARstart{D}{etecting} the first break (FB) of P-wave or S-wave in pre-stack seismic gather is a crucial problem in seismic data processing. Accurate FB picking can provide precise static correction results, significantly improving the performance of subsequent seismic data processing\cite{yilmaz2001seismic}, e.g., velocity analysis, stratigraphic imaging, etc. 
The current inefficient manual FB picking cannot satisfy the requirement because of the increase in the density of seismic data acquisition. Thus, the automatic FB picking algorithm has become a fundamental problem in reducing the seismic processing period.

In the initial exploration of automatic picking methods, many techniques identify FB by analyzing its signal characteristics. These characteristics include energy ratio-based methods\cite{allen1978automatic}\cite{baer1987automatic}\cite{sabbione2010automatic}\cite{gaci2013use}, higher-order statistics-based methods\cite{yung1997example}\cite{Saragiotis2004}, etc. Concretely, taking the feature named the short- and long-time average ratio (STA/LTA)\cite{allen1978automatic}\cite{baer1987automatic} as an example, the local feature of a single trace is calculated by a sliding window. The time at which the STA/LTA value first exceeds the threshold (set in advance) is considered FB. Then, considering the correlation between multiple signals, the correlation coefficient is proposed to identify FB\cite{molyneux1999first}\cite{raymer2008semiautomated}. Later, the feature signal extraction method of multiple signals is proposed one after another\cite{irving2007improving}\cite{de20091998}\cite{qin2021method}\cite{kim2023first}.

Different from the above traditional model-driven methods, the current rapid development of picking methods based on machine learning (ML) or deep learning (DL) is a kind of data-driven approach\cite{hu2019first}\cite{ma2019automatic}\cite{P2021Neurips}\cite{han2021first}. Generally, these data-driven models must train on a few samples with manual labels before predicting FB. In general, current popular methods can be divided into three classes: clustering method-based methods, local image classification-based methods, and end-to-end segmentation-based methods. 
First, the clustering-based picking method generally transforms seismic signal data into other domains, then uses the clustering methods to split the clusters, and finally defines the boundary of two clusters as FB. Concretely, Xu et al. \cite{xu2020high} proposed a multi-stage clustering-based method, which first determines the first-break time window, then utilizes an improved clustering method to obtain the initial FBs of every single trace, and finally considers comprehensively the picking results of the near traces to refine the pickings.
Then, Gao et al. \cite{gao2021stable} considered the clustering methods based on a shot-gather (multiple signals level) to enhance the correlation of automatic pickings among the near traces. Currently, Lan et al. \cite{lan2021automatic} proposed a more complex and robust automatic picking method based on fuzzy C means clustering (FCM) and Akaike information criterion (AIC) further to improve the accuracy and robustness of clustering-based methods.
Second, the local image classification-based methods split the whole shot gather image into a few mini-patch and then utilize neural networks (NN) or machine learning classification methods to classify the mini-patch sub-image into two classes, i.e., FB or non-FB. At a very early time, McCormack et al. \cite{mccormack1993first} proposed using a fully connected network (FCN) to complete the binary classification task. 
Then, FCN was replaced by convolutional neural networks (CNN) for image classification to improve classification accuracy\cite{yuan2018seismic}. Different from classification for picking FB directly, Duan et al. \cite{duan2020multitrace} first obtain a preliminary picking result based on the traditional picking method and then utilize CNN to identify poor picks. Currently, Guo et al. \cite{guo2020aenet} combined more feature information as the input of the classifier and proposed a new post-processing method to improve the stability of the picking results. Compared to NN-based methods, the classic machine learning-based methods outperform in a small sample size scenario, e.g., support vector machine (SVM)\cite{duan2019multi}, support vector regression (SVR), and extreme gradient boosting (XGBoost)\cite{mkezyk2019multi}. Third, end-to-end segmentation-based methods are currently the most popular automatic picking methods since fully convolutional networks (FCN)\cite{long2015fully} and U-Net\cite{ronneberger2015u} are widely used for pixel-level segmentation and have excellent segmentation accuracy. Concretely, FCN was applied to segment the position of FB in the image of a shot gather, and the techniques of semi-supervised learning and transfer learning were conducted to learn a better high-level feature\cite{tsai2018first}\cite{tsai2019automatic}\cite{xie2019first}. Then, U-Net, an excellent medical image segmentation tool, was utilized to solve the segmentation task of labeling FBs\cite{hu2019first}\cite{ma2019automatic}\cite{P2021Neurips}\cite{han2021first}. Yuan et al. \cite{yuan2020robust} further proposed post-processing to refine the U-Net-based picking results, where a recurrent neural network (RNN) regresses FBs for every trace. Moreover, the popular transformer has also been used for the segmentation task concerning picking FB\cite{jiang2023seismic}. 

Semantic segmentation-based methods are the most efficient and have been well-evaluated in practical applications. Unfortunately, DNNs are black-box models, and we cannot explain the specific role of each neural layer\cite{lecun2015deep}. Therefore, controlling the physical significance of each layer output is impossible. However, the principle of the FB picking task is to put quality before quantity, i.e., the accuracy should be guaranteed even if the picking rate is reduced\cite{yilmaz2001seismic}.
To solve the security problem of automatic picking, we introduce the concept of uncertainty quantification (UQ) into FB picking. 
UQ methods mainly measure two parts of uncertainty, epistemic uncertainty (model uncertainty) and aleatoric uncertainty (data uncertainty)\cite{abdar2021review}. 
In this paper, we focus on the epistemic uncertainty of neural networks, where the main task is modeling and solving the posterior distribution of the model weights.
Specifically, the obtained posterior distribution can help us estimate the uncertainty of the FB picked by NN and then filter the outliers among the automatic pickings.
With the proposed methods of Monte Carlo (MC) dropout\cite{gal2016dropout}, Bayes by Backprop (BBB) \cite{blundell2015weight}, and deep ensemble\cite{lakshminarayanan2017simple}, the computational cost of uncertainty quantification is greatly reduced, and the generalization of the model is dramatically improved. Further, uncertainty quantification methods are extended to semantic segmentation\cite{kendall2015bayesian}, especially to the segmentation tasks of medical images\cite{sankaran2015fast}\cite{mehrtash2020confidence}, remote sensing images\cite{kampffmeyer2016semantic}, etc. However, no uncertainty quantification method fully applies to segmentation-based FB picking methods. 

To analyze the uncertainty of the FB pickings of the DL-based method, we propose a novel uncertainty-based picking network named UPNet, which includes a Bayesian segmentation network (BSN), a multi-information regression network (MIRN), and an uncertainty-based decision method (UDM). Concretely, BSN estimates a posterior distribution of the FB picking map of a 2-dimension (2D) gather. MIRN infers the accurate FB time based on the segmentation map sampled from the above posterior distribution. Finally, UDM decides on the final FB picking based on the uncertainty of MIRN regression results.

The structure of this article is summarized as follows. Section II introduces preliminary knowledge about FB picking and uncertainty quantification. 
Section III describes the details of our proposed UPNet. Then, performance validation of the proposed method on four field datasets is shown in Section IV. 
Finally, Section V concludes this article.

\section{Preliminary}

\subsection{Problem Statement}
\begin{figure}[!ht]
    \centering
    \includegraphics[width=0.48\textwidth]{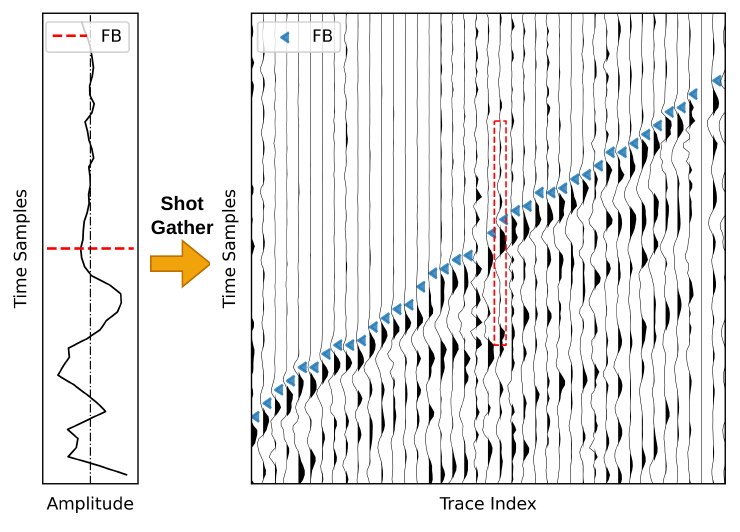}
    \caption{A showcase of the FB picking task. The left subfigure shows the FB of a single trace. The right subfigure indicates a high correlation among the FBs of the adjacent traces. The red dotted box indicates the location of the single-channel signal in the left subfigure.}
    \label{fig: FB-show}
\end{figure}

FB picking is a problem of detecting the first changed point, which indicates a first arrival wave event in the seismic signal data. For instance, the red dash line (Fig.~\ref{fig: FB-show}, left) labels the FB since there is the first pronounced trough. Generally, the source determines the FB mode (peak or trough).
In a higher scale view, the FBs of the adjacent signals on a shot gather usually have a stronger correlation, as shown in the right subfigure of Fig.~\ref{fig: FB-show}.
Based on the spatial characteristics of FB mentioned above, we define the FB picking task as a statistical inference problem $P(\mathbf{t}|\mathbf{G})$, which infers the FB of a time series $\mathbf{t}$ based on the shot gather $\mathbf{G}$. 

\subsection{Uncertainty quantification for FB Picking}
We transform the picking model to a probability model to introduce the uncertainty of FB.
In view of statistic, the inference of FB $P(\mathbf{t}|\mathbf{G})$ can be described as solving a maximum posterior estimation:
\begin{equation}
    \mathbf{\hat{t}} = \arg\max_{\mathbf{t}}\{P(\mathbf{t}|\mathbf{G}, \mathcal{D})\},
    \label{MAP}
\end{equation}
where $\mathbf{G}$ represents a shot gather, $\mathbf{t}$ represents the FB of each trace in the shot gather $\mathbf{G}$, and $\mathcal{D}$ represents the annotation knowledge of FB, i.e., the labeled dataset. In this paper, we only consider the model uncertainty, i.e., epistemic uncertainty, in the inference model (Eq. \ref{MAP}). Concretely, the model weights $\mathbf{w}$ are assumed to follow a posterior distribution $P(\mathbf{w}|\mathcal{D})$. Therefore, the posterior distribution can be expressed by a form of marginal average:
\begin{equation}
    P(\mathbf{t}|\mathbf{G}, \mathcal{D}) = \int_{\mathbf{w}}{P(\mathbf{t}|\mathbf{G},\mathbf{w})\cdot P(\mathbf{w}|\mathcal{D}) d\mathbf{w}},
    \label{MargAvg}
\end{equation}
where $P(\mathbf{t}|\mathbf{G},\mathbf{w})$ is an inference model with weights $\mathbf{w}$.

\begin{figure*}[!ht]
    \centering
    \includegraphics[width=\textwidth]{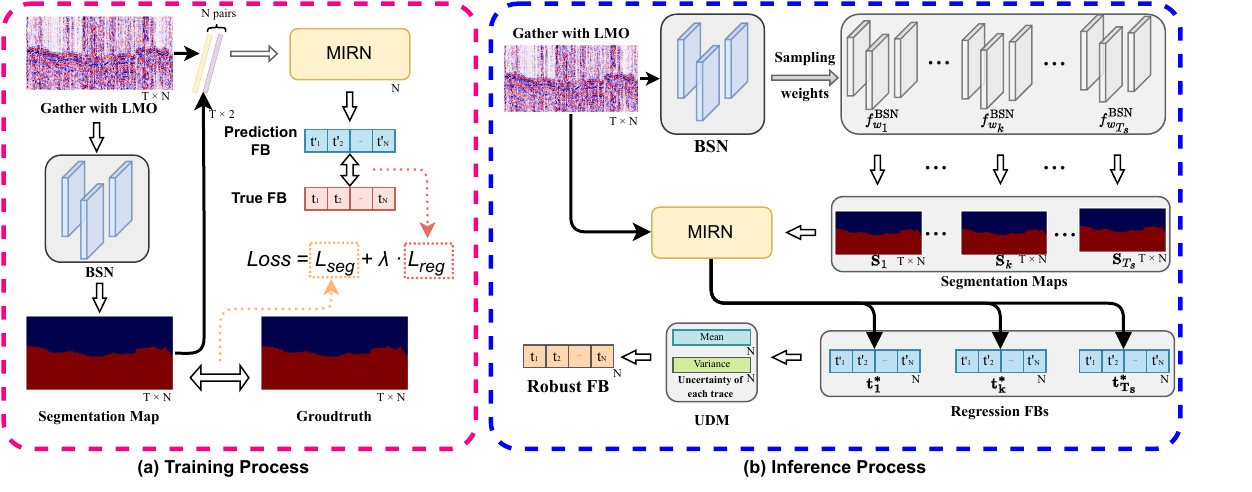}
    \caption{The main flow chart of a new framework we propose to pick FB times robustly.}
    \label{fig: mainflow}
\end{figure*}

However, the true posterior $P(\mathbf{w}|\mathcal{D})$ is intractable, and we have to utilize other techniques to approximate it, e.g., variational inference (VI)\cite{fox2012tutorial}, which minimizes the Kullback-Leibler (KL) divergence between the true posterior $P(\mathbf{w}|\mathcal{D})$ and an approximation distribution $q(\mathbf{w})$:
\begin{equation}
    KL\left[q(\mathbf{w})||P(\mathbf{w}|\mathcal{D})\right] = \int q(\mathbf{w})\ln \frac{q(\mathbf{w})}{P(\mathbf{w}|\mathcal{D})}d\mathbf{w},
    \label{KL}
\end{equation}
Then, we maximize the evidence low bound (ELBO) deduced by Eq. \ref{KL} to optimize the approximation distribution $q(\mathbf{w})$:
\begin{equation}
    \text{ELBO}(q) = KL\left[q(\mathbf{w})||P(\mathbf{w},\mathcal{D})\right] - \ln P(\mathcal{D}).
    \label{ELBO}
\end{equation}
However, traditional VI-based methods could be more inefficient in optimizing Eq.~\ref{ELBO} and are hard to apply to deep learning-based methods directly. Fortunately, Gal and Ghahramani\cite{gal2016dropout} claimed that Monte Carlo dropout applied in a deep neural network can be approximated to a VI processing. The dropout technique is usually used to avoid the overfitting performance of training processing and is only conducted in training processing. Unlike only using dropout in training processing, \cite{gal2016dropout} enables the neural network model to a posterior probability model with the help of dropout layers.
Concretely, the dropout technique defines a series of Bernoulli variables $z$ and multiplies it by the weight matrix $\mathbf{M}$ of a full-connection or convolutional layer, e.g., $\mathbf{M_i}$ is the weight matrix of the $i$th layer, that is,
\begin{equation}
    \mathbf{W_i} = \mathbf{M_i} \cdot \text{diag}(\left[z_{i,j}\right]_{j=1}^{K_i}),
    \label{dropout}
\end{equation}
\begin{equation}
    z_{i,j} \thicksim Bernoulli(p_i), i=1,...,L, j=1,..,K_{i-1},
    \label{binary}
\end{equation}
where $p_i$ is the dropout rate of $i$th layer, $L$ is the number of layers, and $K_i$ is the number of neural units of $i$-th layer. 
Further, Gal and Ghahramani\cite{gal2016dropout} proved that NN with the dropout technique is a posterior probability model. 
Next, we inspect the posterior distribution of the model output $\mathbf{t^*} \thicksim q(\mathbf{t^*}|\mathbf{G^*}, 
\mathcal{D})$. 
Specifically, we sample the model weights $T_s$ times from $q(\mathbf{w}|\mathcal{D})$, denoted as $\mathbf{w^{(1)}}, ..., \mathbf{w^{(T_s)}}$, and infer FB based on $T_s$ models respectively:
\begin{equation}
\mathbf{\hat{t}_k}=f_{\mathbf{w^{(k)}}}(\mathbf{G^*}), k=1,...,T_s,
    \label{MC-sampling}
\end{equation}
where $\mathbf{\hat{t}_k}$ is the predicted FB vector using the $k$-th sampled model. The above sampling is sampling $z$ in Eq.~\ref{binary}.
The statistical characteristics of the posterior distribution $q(\mathbf{t^*}|\mathbf{G^*}, \mathcal{D})$, e.g., the mean and the variance, are estimated by Monte Carlo methods. Thus, Gal and Ghahramani\cite{gal2016dropout} noted this type of dropout as the MC dropout. First, the mean can be inferred by the average of sampled FBs:
\begin{equation}
       \text{Mean}({t^*_j}) = E_q\left[{t^*_j}\right]  \approx \frac{1}{T_s}\sum\limits_{k=1}^{T_s}{\hat{t}_{j,k}},
    \label{MC-mean}
\end{equation}
where ${t^*_j}$ represents the random variable of $j$-th FB, $\hat{t}_{j,k}$ is the $j$-th element of $k$-th sampled prediction (Eq.~\ref{MC-sampling}), and $E_q\left[\cdot\right]$ means the expectation based on the posterior distribution $q(\mathbf{t^*}|\mathbf{G^*}, \mathcal{D})$.
Momentously, the mean of $\mathbf{{t^*}}$ usually is a robust prediction of Bayesian NN, and we also take it as the final output of NN in this paper.
More importantly, the uncertainty of FB can be captured by estimating the variance of $\mathbf{t^*}$ using the Monte Carlo method as well.
Concretely, the variance of $j$-th element of $\mathbf{t^*}$ is estimated by:
\begin{equation}
       \text{Var}({t_j^*}) = E_q\left[{t_j^*}^2\right] - \left[E_q{t_j^*}\right]^2 \approx \frac{1}{T_s}\sum\limits_{k=1}^{T_s} {\hat{t}_{j,k}}^2 - \left[\frac{1}{T_s}\sum\limits_{k=1}^{T_s}{{\hat{t}_{j,k}}}\right]^2.
    \label{MC-var}
\end{equation}
Overall, the estimated mean can be used for FB prediction, and the computed variance can help determine trust in FB.

\section{Methodology}
UPNet consists of two processes: a training process and an inference process, as illustrated in Fig. \ref{fig: mainflow}. Since the Bayesian Segmentation Network (BSN) and the Multi-Information Regression Network (MIRN) are two neural networks (NNs), the training process optimizes the weights of BSN and MIRN using a joint-loss function as shown in Fig. \ref{fig: mainflow}(a). After the training process, UPNet infers robust FB picking through BSN, MIRN, and the Uncertainty-based Decision Method (UDM) sequentially. Next, we introduce the principles of each component and the details of the training process and inference process in turn.

\subsection{Bayesian Segmentation Network (BSN)}
\begin{figure}[!ht]
    \centering
    \includegraphics[width=3.3in]{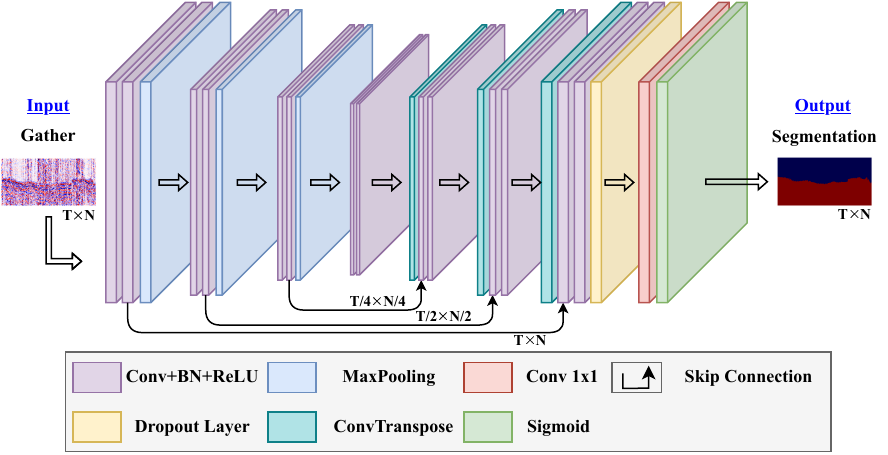}
    \caption{BSN: a U-Net with the MC dropout technique. The input of BSN is a 2-dimensional (2D) gather, and the output is a segmentation map that indicates the region before FB (pixels labeled 0, blue) and the region after FB (pixels labeled 1, red).}
    \label{fig: BUNet}
\end{figure}

To represent the distribution of the predicted FB, we construct a DL-based probability model, namely, BSN. Specifically, we utilize a popular semantic segmentation model, the U-Net\cite{ronneberger2015u}, to detect the signal of FB and employ the Monte Carlo (MC) dropout technique\cite{gal2016dropout} to convert the U-Net into a probabilistic model. Fig. \ref{fig: BUNet} illustrates the structural design of BSN. The input of BSN is a shot gather, and the output is the binary segmentation map. This map labels the noise before FB as 0 and the signal after FB as 1. Next, we will introduce the details of BSN.

The U-Net we use is a fully convolutional neural network designed with an encoder-decoder architecture and skip connections. In this paper, the depth is set to 3, indicating that down-sampling is performed three times.
The convolutional (Conv) layers, Batch-Normalization (BN) layers, rectified linear units (ReLU)  activation layers\cite{nair2010rectified}, Max-Pooling layers, dropout layers, transposed convolution (ConvTranspose) layers, and Sigmoid layer are the main components of the BSN. The convolutional unit is repeated many times (purple layers in Fig.~\ref{fig: BUNet}) in the entire processing, and the model includes a Conv layer (3$\times$3 kernel, stride = 1, and padding = 1), a BN layer, and a ReLU activation layer. The dropout layer with a constant dropout rate is applied after the final convolutional unit. In the encoder section, Max-Pooling layers (with a stride of 2) are utilized to downsample and compress redundant information. In the decoder section, we use ConvTranspose layers for up-sampling, with a kernel size = 2$\times$2 and a stride of 2. The skip connections are also used to concatenate the decoded features and the original shallow encoded features, aiming to recover the segmentation details better. Finally, we use a Conv layer with a 1$\times$1 kernel to compress and summarize the decoded outputs without changing the output shape. We also employ a Sigmoid layer to normalize the segmentation map within the $\left[0, 1\right]$ range.
Moreover, the kernel numbers of each Conv layer should be underlined to equal the number of output channels. There are 15 Conv layers with kernel numbers of 32, 32, 64, 64, 128, 128, 256, 256, 128, 128, 64, 64, 32, 32, and 1, respectively.

Since BSN is a probability model, the model weights should be sampled in the inference process from the posterior distribution of weights $w \backsim p(w|\mathcal{D})$ (in Eq.~\ref{MargAvg}). The sampling processing is defined as the Bernoulli variables of dropout layers sampled from a Bernoulli distribution with a successful probability equal to the dropout rate. Then, the BSN with sampled weights infers the segmentation map of FB $\mathbf{S}$ for the gather $\mathbf{G}$:
\begin{equation}
    \mathbf{S} = f^{\text{BSN}}_{w}(\mathbf{G}),
 \label{BSN}
\end{equation}
where $f^{\text{BSN}}_{w}$ is the BSN with the sampled model weight $w$. 
To approximate the distribution of the segmentation map of the gather $\mathbf{G}$, we conduct sampling $T_s$ times and obtain corresponding $T_s$ group weights.
Accordingly, we can calculate the $T_s$ segmentation graph, denoted as $\mathbf{S}_1, \ldots, \mathbf{S}_{T_s}$.

\subsection{Multi-Information Regression Network (MIRN)}
\begin{figure}[!ht]
    \centering
    \includegraphics[width=3.3in]{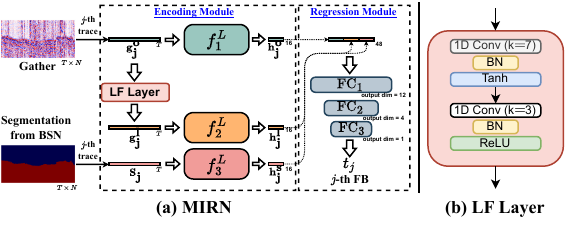}
    \caption{The flow chart of MIRN. A 2D gather is divided into N traces, and MIRN picks FB trace by trace. The shape (or length) of each element is marked in the lower right corner.}
    \label{fig: MIRN}
\end{figure}

Above BSN has given a FB segmentation on a large scale. However, there are still pixel-scale deviations if we use the threshold method to obtain FB from the segmentation map\cite{hu2019first}. Thus, MIRN is proposed to correct the errors using a regression strategy. 
In Fig. \ref{fig: MIRN}(a), MIRN includes an encoding module and a regression module. On one hand, the encoding module encodes three parts of information: the segmentation map of BSN, the original signal of traces, and the low-frequency signal of the traces. We claim that BSN has analyzed the FB region on a global level. Thus, it is sufficient for MIRN to infer specific FB values trace by trace. 
The segmentation map $\mathbf{S}$ (Eq. \ref{BSN}) is divided by columns into $N$ vectors, $\{\mathbf{s_j}\}_{j=1}^N$, $j$ from 1 to $N$, where $N$ represents the number of traces in a gather. For convenience, we take the calculation of $j$-th trace as an example. The original signal of $j$-th trace is called $\mathbf{g^o_j}$. To capture the local feature, we also encode the low-frequency signal of $\mathbf{g^o_j}$, denoted as $\mathbf{g^l_j}$. 
Fig. \ref{fig: MIRN}(b) illustrates the specific design of the low-frequency signal convolutional layer. Concretely, there are two Conv modules, which contain six layers in total. The first Conv module includes a 1D Conv layer with a kernel length of 7, striding of 1, and padding of 3, followed by a BN layer and a hyperbolic tangent (Tanh) activation layer\cite{lecun1998gradient}, sequentially. The second Conv module consists of a 1D Conv layer with a kernel size of 3, stride of 1, and padding of 1. A BN layer and a ReLU activation layer are in sequence, followed by this. Mainly, since the value range of $\{\mathbf{g^o_j}\}_{j=1}^N$ is [-1, 1], a Tanh activation layer is applied in the first layer to enhance the ability to capture phase information. Subsequently, three Long Short-Term Memory (LSTM) layers\cite{gers2000learning} $f^L_1, f^L_2, f^L_3$ encode $\mathbf{g^o_j}, \mathbf{g^l_j}, \mathbf{s_j}$ into three hidden vectors $\mathbf{h^o_j}$, $\mathbf{h^l_j}$, and $\mathbf{h^s_j}$, respectively, where the length of all three hidden vectors is 16. To match the complexity of the model and the data, we set different numbers of recurrent layers for $f^L_1, f^L_2$, and $f^L_3$, i.e., three, two, and one. 
On the other hand, the regression module fuses three encoded vectors ($\mathbf{h^o_j}$, $\mathbf{h^l_j}$, and $\mathbf{h^s_j}$). It maps the fused vector to the corresponding pixel index of FB using three full connection (FC) layers as shown in the Regression Module of Fig. \ref{fig: MIRN}. Specifically, $\mathbf{h^o_j}$, $\mathbf{h^l_j}$, and $\mathbf{h^s_j}$ are sequentially concatenated into a vector of length 48. Subsequently, three FC layers output two vectors of length 12 and 4 and a scale value, respectively. Particularly, the value range of the scale value is [0, 1], which represents the percentage of the relative position in the original sampling length $T$ in the time domain. Thus, we multiply this scalar by the length $T$ to recover the corresponding index of FB. In MIRN, we repeat this process for each trace and obtain the FB indexes, denoted as $\mathbf{t^*}=[\hat{t}_1, ..., \hat{t}_N]$.

\subsection{Uncertainty-based Decision Method (UDM)}
UDM consists of two steps: calculating uncertainty and removing unstable picking. On the one hand, MIRN can regress corresponding FB vectors based on the segmentation maps sampled from BSN, and we denoted them as $\mathbf{t^*_{(1)}}, ..., \mathbf{t^*_{(T_s)}}$, where $T_s$ is sampling times of BSN and the length of each vector is $N$. Thus, the uncertainty of FB corresponding to $j$-th trace can be expressed by the variance of $T_s$ pickings (Eq. \ref{MC-var}):
\begin{equation}
    \text{Var}(t^*_j) \approx \frac{1}{T_s}\sum\limits_{k=1}^{T_s} {\hat{t}_{j,k}}^2 - \left[\frac{1}{T_s}\sum\limits_{k=1}^{T_s}{{\hat{t}_{j,k}}}\right]^2,
 \label{UDM-var}
\end{equation}
where $\hat{t}_{j,k}$ represents $j$-th element in $\mathbf{t^*_{(k)}}$, where $j=1,...,N$.
On the other hand, unstable picking can be removed based on the estimated variance. Since variance is a random variable, we can calculate an empirical distribution of variance. Subsequently, we obtain a threshold of variance $t_{\text{var}}(p)$, which ensures:
\begin{equation}
    \left|\frac{\#(\hat{F}_v(v)<t_{\text{var}}(p))}{N_{\text{all}}} - p\right|<0.1\%,
 \label{UDM-thre}
\end{equation}
where $\hat{F}_v$ represents the empirical distribution function (EDF) of the variance, $\#(\cdot)$ represents the count function, $v$ represents the random variable corresponding to variance, and $N_{\text{all}}$ is the number of all traces. 
We call $p$ confidence level, which can be determined by the quality of the survey.
For instance, $p$ can be set as 0.9 when the survey data has high SNR and 0.6 when the survey data has low SNR. Subsequently, the pickings with the variances greater than $t_{\text{var}}(p)$ are removed. Finally, the FBs of the remaining traces with low picking variances are determined by calculating the mean value of the sampled picks:
\begin{equation}
    \hat{t}_j = E(t^*_j) \approx \frac{1}{T_s}\sum\limits_{k=1}^{T_s} {\hat{t}_{j,k}},
\label{UDM-mean}
\end{equation}
where $j$ is in the index set of the remaining traces.

\subsection{Training and Inference of UPNet}
\subsubsection{Preprocess}
Before starting training and inference, the traces in the 2D gather should be preprocessed, including linear moveout (LMO) correction and trace-wise normalization. On the one hand, we crop the input region of the gather based on the LMO correction, which utilizes a velocity prior to the current survey to estimate reference arrival times for each trace. Especially, the reference arrival time of the $j$-th trace $t_j$ is computed by:
\begin{equation}
  t_j = d_j / v + t_0
  \label{LMO}
\end{equation}
where $d_j$ is the offset, i.e., the distance between the source and the receiver. $v$ and $t_0$ are the reference velocity and the interception time of the prior information in the current survey, which is detailed in III.B. Therefore, the top boundary $\text{U}_j$ and the bottom boundary $\text{D}_j$ of the crop index of $j$-th trace are computed by: 
\begin{equation}
    \text{U}_j = t_j - \frac{h}{2}; \text{D}_j = t_j + \frac{h}{2} - 1,
    \label{LMO_crop}
  \end{equation}
where $h$ represents the cropped length. 
On the other hand, we normalize each trace of the gather after LMO correction using the trace-wise normalization method:
\begin{equation}
    g^n_{ij} = \frac{g_{ij} }{\max\limits_{i={1,...,T}}\{\left|g_{ij}\right|\}}, j=1,...,N,
    \label{TW-norm}
  \end{equation}
where $g_{ij}$ represents the amplitude of the $i$-th row and $j$-th column of the gather $\mathbf{G}$, $g^n_{ij}$ is the corresponding normalization value, and $N$ is the trace number of the gather $\mathbf{G}$. After these two steps, we crop the gather to a smaller shape and map it into the range of [-1, 1].

\subsubsection{Training Process}
UPNet contains two learning tasks: a segmentation task (BSN) and a regression task (MIRN). To improve the cooperative picking ability of BSN and MIRN, we train them simultaneously in the same training framework as shown in Fig. \ref{fig: mainflow}(a). However, the optimization problem of the multi-task model is intractable. We must keep the balance among the loss function of various tasks to mitigate the disappearance of gradients. In our study, we adopt a balance strategy of loss functions, named MetaBalance\cite{he2022metabalance}:
\begin{equation}
    \text{L}^{\text{B}} = \text{L}^{\text{Seg}} + \text{L}^{\text{Reg}} \cdot \frac{\left|\text{L}^{\text{Reg}}\right|}{\left|\text{L}^{\text{Seg}}+\epsilon\right|},
    \label{MetaBalance}
  \end{equation}
where $\text{L}^{B}$, $\text{L}^{\text{Seg}}$ and $\text{L}^{\text{Reg}}$ are the loss function of the balanced, the segmentation task, and the regression task, respectively. Especially, the adaptive balance parameter $\frac{\left|\text{L}^{\text{Reg}}\right|}{\left|\text{L}^{\text{Seg}}+\epsilon\right|}$ does not participate in the backward propagation of the gradient, and $\epsilon$ is a positive number very close to 0 to ensure division. Specifically, the loss function $L^{\text{Seg}}$ used in the segmentation task is the binary cross entropy (BCE) loss function to supervise the difference between the segmentation maps of ground truth $\mathbf{S^{\text{GT}}}$ and prediction $\mathbf{S^{\text{P}}}$. The specific loss function on a pair of $\mathbf{S^{\text{GT}}}$ and $\mathbf{S^{\text{P}}}$ is defined by:
\begin{equation}
    L^{\text{Seg}}(\mathbf{S^{\text{P}}}, \mathbf{S^{\text{GT}}}) = - \sum\limits_{i=1}^{T}\sum\limits_{j=1}^{N}\frac{s^{\text{GT}}_{ij} \cdot \log s^{\text{P}}_{ij} + (1 - s^{\text{GT}}_{ij}) \cdot \log (1 - s^{\text{P}}_{ij})}{T\cdot N},
    \label{BCELoss}
  \end{equation}
where $s^{\text{GT}}_{ij}$ and $\log (1 - s^{\text{P}}_{ij})$ are the $i$-th row and $j$-th column values of the $\mathbf{S^{\text{GT}}}$ and $\mathbf{S^{\text{P}}}$, respectively, and $T$ and $N$ are the number of time samples and traces, respectively. Moreover, we use the smooth L1 loss function to supervise the deviation between manual FB $\mathbf{t^{M}}$ and automatic FB $\mathbf{t^{A}}$. The smooth L1 loss function defined on a gather is: 
\begin{equation}
    \begin{array}{c}
    \text{L}^{\text{Reg}} (\mathbf{t^{A}}, \mathbf{t^{M}}) = \sum\limits_{j=1}^{N}{\frac{l_j}{N}}\\
    l_j = \begin{cases}
        0.5 (t^M_j - t^A_j)^2 / \beta, & \text{if } |t^M_j - t^A_j| < \beta \\
        |t^M_j - t^A_j| - 0.5 \times \beta, & \text{otherwise},
        \end{cases}    
    \end{array}
    \label{SL1Loss}
  \end{equation}
where $t^M_j$ and $t^A_j$ are the $j$-th values of $\mathbf{t^{A}}$ and $\mathbf{t^{M}}$.

\subsubsection{Inference Process}
After the training process, UPNet infers the FB of a new 2D gather that has not been seen before through four sequential steps as shown in Fig. \ref{fig: mainflow}. First, we sample $T_s$ groups of the weights, denoting these BSNs as $f^{\text{BSN}}_1, \ldots, f^{\text{BSN}}_{T_s}$. Second, the preprocessed gather is input into these BSNs, and the segmentation maps $\mathbf{S_1}, \ldots, \mathbf{S_{T_s}}$ are computed. 
Third, the trained MIRN predicts the corresponding FBs $\mathbf{t^*_1}, \ldots, \mathbf{t^*_{T_s}}$ for $\mathbf{S_1}, \ldots, \mathbf{S_{T_s}}$. 
Finally, the sampled FBs $\mathbf{t^*_1}, \ldots, \mathbf{t^*_{T_s}}$ are filtered by the UDM, and we obtain the inferred FBs $\{\hat{t}_j\}_{j=1}^N$ for $N$ traces. 
Since the inference processes of the BSNs are independent, the computation processes can be parallel. Furthermore, the same applies to MIRN. Thus, even though the sampling time is extensive, the computational time of UPNet can be managed.

\section{Experiments}


\subsection{Datasets and Metrics}
To facilitate comparison of other popular methods and other researchers expediently reproduce our approach, we test on a group of the open source field datasets from all unique mining sites in three provinces of Canada provided by \cite{st2024deep}, including four datasets, named Lalor, Brunswick, Halfmile, and Sudbury, respectively. More details of each dataset are shown in Tab.~\ref{Tab: datasets}, in which label ratio is the ratio of the number of labeled traces to the number of all traces. The provided FB labels are first obtained using the STA/LTA method and then corrected by experts. Notably, each FB in these datasets is annotated at the trough of the arrival waveform. 

\begin{table}[!ht]
\centering
\caption{Open dataset information}
\resizebox{\linewidth}{!}{\begin{tabular}{lcccc}
\toprule
Survey        & Shot Gather Num. & Annotation Ratios & Sample Interval & Sample Times \\ \hline
Brunswick & 18475            & 83.54\%              & 2ms   & 751       \\
Lalor    & 14455            & 46.13\%              & 1ms   & 1001      \\
Sudbury    & 11420            & 12.9\%               & 2ms   & 751       \\
Halfmile  & 5520             & 90.35\%              & 2ms   & 1001      \\ \bottomrule
\end{tabular}}\label{Tab: datasets}
\end{table}

To evaluate the FB of automatically picking methods, we adapt the metrics of a benchmark method of the open-source dataset we use\cite{st2024deep}. For convenience, $\mathbf{t^{\text{A}}}$ and $\mathbf{t^{\text{M}}}$ represent the picking vector of automatical methods and humans on a 2D gather. Particularly, both lengths of them are $N$, and the FB values of the traces removed by UDM are -1. Since UPNet removes some picks with low confidence, we introduce a new metric called automatic pick rate (APR), not mentioned in \cite{st2024deep}, which calculates the rate of automatic picks compared to manual picks:
\begin{equation}
    \text{APR}(\mathbf{t^{\text{A}}}) = \frac{1}{N}{\sum\limits_{j=1}^{N}I\{t^{\text{A}}_j>0\}}
     \label{APR}
 \end{equation}
where $t^{\text{A}}_j$ is the $j$-th element of $\mathbf{t^{\text{A}}}$, and $I\{\cdot\}$ is the indicator function. Next, we will briefly introduce the metrics utilized in the study\cite{st2024deep}. 
First, the hit rate within k pixel (HR@$k$px, $k=1, 3, 5, 7$, and $9$) is used to measure the accuracy of automatic FB:
\begin{equation}
    \text{HR}@k\text{px}(\mathbf{t^{\text{A}}}, \mathbf{t^{\text{M}}}) = \frac{\sum\limits_{j=1}^{N}I\{\left|t^{\text{A}}_j-t^{\text{M}}_j\right|<k\}\cdot I_j}{\sum\limits_{j=1}^{N}{I_j}},
    \label{HR}
 \end{equation}
where $t^{\text{M}}_j$ is the $j$-th element of $\mathbf{t^{\text{M}}}$, and $$I_j= I\{t^{\text{A}}_j>0, t^{\text{M}}_j>0\}.$$ Second, the mean absolute error (MAE) measures the mean deviation between the automatic picking and the manual picking:
\begin{equation}
   \text{MAE}(\mathbf{t^{\text{A}}}, \mathbf{t^{\text{M}}}) = \frac{\sum\limits_{j=1}^{N}\left|t^{\text{A}}_j-t^{\text{M}}_j\right|\cdot I_j}{\sum\limits_{j=1}^{N}{I_j}}.
    \label{MAE}
\end{equation}
Third, the root mean square error (RMSE) represents the stability of the automatic picking and is defined by:
\begin{equation}
    \text{RMSE}(\mathbf{t^{\text{A}}}, \mathbf{t^{\text{M}}}) = \frac{\left(\sum\limits_{j=1}^{N}\left|t^{\text{A}}_j-t^{\text{M}}_j\right|^2\cdot I_j\right)^{1/2}}{\sum\limits_{j=1}^{N}{I_j}}.
     \label{RMSE}
 \end{equation}
Finally, the mean bias error (MEB) is defined by:
\begin{equation}
    \text{MBE}(\mathbf{t^{\text{A}}}, \mathbf{t^{\text{M}}}) = \frac{\sum\limits_{j=1}^{N}\left(t^{\text{A}}_j-t^{\text{M}}_j\right)\cdot I_j}{\sum\limits_{j=1}^{N}{I_j}}.
     \label{MBE}
 \end{equation}
To some extent, MBE can represent the preference of automatic picking methods for position picking. For instance, MBE $<$ 0 indicates that the method favors selecting the pixels before those corresponding to FBs.

\subsection{Implement Details}
This subsection has three aspects: a four-fold experiment setting, hyperparameters setting of the LMO correction, and training hyperparameters tuning method. 
First, to compare with the benchmark method\cite{st2024deep}, we followed the same four-fold experiment described in their study, where the model is evaluated in four dataset splits, respectively. Tab. \ref{Tab: fold_set} shows four training scenarios that mimic real-world applications. This involves training on a few field surveys and predicting results for other surveys that have not been seen before. To avoid experimental variability, these four folds use different combinations of training sets and are tested on various validation and test sets. Moreover, the experiments of each fold are independent of each other, so data leakage cannot occur.
\begin{table}[!ht]
    \centering
    \caption{Four-Fold Experiment Setting}
    \resizebox{0.8\linewidth}{!}{\begin{tabular}{llll}
        \toprule
        \textbf{Fold} & \textbf{Training Sites} & \textbf{Validation Sites} & \textbf{Test Sites} \\ \hline
        1             & Halfmile, Lalor         & Brunswick                 & Sudbury             \\
        2             & Sudbury, Halfmile       & Lalor                     & Brunswick           \\
        3             & Lalor, Brunswick        & Sudbury                   & Halfmile            \\
        4             & Brunswick, Sudbury      & Halfmile                  & Lalor               \\ \bottomrule
        \end{tabular}}\label{Tab: fold_set}
    \end{table}
Second, the LMO correction has two important hyperparameters (Eq. \ref{LMO}): the reference velocity and the interception time $t_0$. Tab. \ref{Tab: LMO} illustrates the details of the hyperparameters used in these four field surveys, which are estimated by a few manually picking FBs. Concretely, the reference velocity ranges from 5136.20 m/s to 6013.97 m/s, and the interception time falls within the range of [0.0017s, 0.0367s]. Subsequently, the LMO correction under these parameters is conducted for each fold, and the height of the cropped gather is 128, i.e., $h=128$ in Eq. \ref{LMO_crop}. Moreover, to avoid the influences of the unpicked region, we only retain the traces with manual annotations and divide the entire gather after LMO correction into pieces with a width of 256. Thus, the final gathers used in experiments are 918, 14582, 3879, and 4389 gathers of shape = (128, 256) in Sudbury, Brunswick, Halfmile, and Lalor, respectively.

\begin{table}[!ht]
    \centering
    \caption{Hyperparameters In the LMO Correction}
    \resizebox{0.8\linewidth}{!}{\begin{tabular}{lcc}
    \toprule
    \textbf{Dataset}   & \textbf{Global Velocity $v$ (m/s)} & \textbf{Interpolation Time $t_0$ (s)} \\ \hline
    Sudbury   & 5891.69                   & 0.0367                       \\
    Brunswick & 5136.20                   & 0.0017                       \\
    Halfmile  & 5349.67                   & 0.0234                       \\
    Lalor     & 6013.97                   & 0.0130                       \\ \bottomrule
    \end{tabular}}\label{Tab: LMO}
    \end{table}

Finally, we tune the training hyperparameters using a grid search method. Specifically, we consider the hyperparameters, including the size of the training mini-batch ($\text{BS}\in\{2, 4, 8, 16, 32, 64, 128\}$), the learning rate of the optimizer ($\text{Lr}\in\{1\text{e}-2, 1\text{e}-3, 1\text{e}-4\}$), optimization method ($\text{Opt}\in\{\text{Adam}$\cite{kingma2014adam}$, \text{SGD}$\cite{sutskever2013importance}$\}$), and the dropout rate ($\text{dr}\in\{0.1, 0.2, 0.3, 0.4, 0.5\}$). Each combination of hyperparameters is repeated ten times using ten random initial seeds of weights. The performance of combinations is evaluated on the mean values of MAE of validation sets over ten training with various random seeds. Tab.~\ref{Tab: tuning} indicates the best combination of hyperparameters for each fold in the grid search. 

\begin{table}[!ht]
\centering
\caption{Hyperparameter Setting of Training Processing}
\resizebox{0.5\linewidth}{!}{\begin{tabular}{lcccc}
    \toprule
    \textbf{Fold} & \textbf{BS} & \textbf{Lr} & \textbf{Opt} & \textbf{dr} \\ \hline
    1             & 32          & 1e-03    & Adam         & 0.1         \\
    2             & 4           & 1e-03    & Adam         & 0.1         \\
    3             & 4           & 1e-03    & Adam         & 0.3         \\
    4             & 4           & 1e-03    & Adam         & 0.1         \\ \bottomrule
    \end{tabular}}\label{Tab: tuning}
\end{table}

\begin{table*}[!ht]
    \center
    \caption{Four-Fold Experimental Results of Five Comparative Methods}
\resizebox{0.8\linewidth}{!}{
    \begin{tabular}{llcccccccc}
        \toprule
        Test Site & Methods              & HR@1px$\uparrow$            & HR@3px$\uparrow$            & HR@5px$\uparrow$            & HR@7px$\uparrow$             & HR@9px$\uparrow$             & RMSE$\downarrow$        & MAE$\downarrow$              & MBE        \\ \hline
        Sudbury   & STA/LTA              & 38                & 51.9              & 79.8              & 81.1               & 81.9               & 15.7             & 7.4              & -4.2       \\
                    & Benchmark            & 73.1±0.5          & 93.9±0.6          & 96.2±0.6          & 97.5±0.5           & 98.2±0.5           & 35.1±12.3        & 2.8±1.2          & 1.2±1.1    \\
                    & Self-trained Network & 74.6±0.1          & 95.0±0.1          & \textbf{97.0±0.1} & \textbf{98.0±0.1}  & \textbf{98.6±0.1}  & \textbf{2.9±0.1} & \textbf{0.6±0.0} & 0.3±0.0    \\
                    & CNNRNN               & 90.0±1.1          & 92.8±0.2          & 94.7±0.1          & 96.0±0.1           & 96.6±0.1           & 6.2±0.1          & 1.3±0.1          & 0.2±0.2    \\
                    & UPNet (Ours)         & \textbf{92.5±0.5} & \textbf{96.0±0.1} & 96.9±0.1          & 97.5±0.0           & 97.8±0.0           & 5.1±0.0          & 1.1±0.0          & 0.1±0.0    \\
        Brunswick & STA/LTA              & 54.3              & 76                & 89.6              & 94.4               & 94.9               & 8.4              & 2.8              & -0.6       \\
                    & Benchmark            & 87.6±1.4          & 96.4±0.6          & 97.8±0.6          & 98.3±0.6           & 98.6±0.6           & 50.2±13.8        & 4.5±2.3          & 3.8±2.5    \\
                    & Self-trained Network & 89.4±0.1          & 96.0±0.1          & 97.5±0.1          & 98.1±0.1           & 98.6±0.1           & 3.1±1.6          & 0.5±0.1          & 0.1±0.1    \\
                    & CNNRNN               & 94.0±0.8          & 97.2±0.3          & 98.3±0.3          & 98.7±0.3           & 99.0±0.3           & 3.2±1.1          & 0.6±0.2          & -0.1±0.2   \\
                    & UPNet (Ours)         & \textbf{99.0±0.2} & \textbf{99.8±0.0} & \textbf{99.9±0.0} & \textbf{100.0±0.0} & \textbf{100.0±0.0} & \textbf{0.4±0.0} & \textbf{0.2±0.0} & -0.1±0.1   \\
        Halfmile  & STA/LTA              & 48.9              & 70.6              & 80                & 90.6               & 92.3               & 9.8              & 3.7              & -0.6       \\
                    & Benchmark            & 83.8±0.5          & 92.6±0.5          & 95.9±0.6          & 97.9±0.6           & 98.8±0.6           & 35.2±33.3        & 3.8±4.3          & 2.9±4.1    \\
                    & Self-trained Network & 84.4±0.3          & 92.6±0.1          & 96.0±0.1          & 98.1±0.1           & 99.2±0.0           & 2.0±0.2          & 0.5±0.0          & 0.0±0.0    \\
                    & CNNRNN               & 88.3±1.6          & 93.5±0.3          & 96.6±0.1          & 98.6±0.1           & 99.4±0.0           & 1.7±0.1          & 0.6±0.0          & 0.1±0.1    \\
                    & UPNet (Ours)         & \textbf{96.1±0.1} & \textbf{96.8±0.1} & \textbf{98.0±0.1} & \textbf{99.2±0.0}  & \textbf{99.7±0.0}  & \textbf{1.2±0.1} & \textbf{0.3±0.0} & -0.0±0.0   \\
        Lalor     & STA/LTA              & 53.5              & 86.2              & 89.5              & 92.9               & 95                 & 6                & 1.9              & 0.2        \\
                    & Benchmark            & 76.3±1.8          & 80.0±1.7          & 82.7±1.7          & 86.4±1.9           & 89.0±2.0           & 460.0±74.0       & 124.0±40.0       & 123.0±40.0 \\
                    & Self-trained Network & \textbf{86.5±0.1} & 88.0±0.1          & 90.5±0.1          & 94.2±0.1           & 96.7±0.1           & 4.6±0.1          & 1.0±0.0          & 0.6±0.0    \\
                    & CNNRNN               & 69.1±5.6          & 80.3±2.3          & 86.1±1.8          & 90.6±1.5           & 93.6±1.3           & 5.1±0.7          & 2.1±0.4          & 0.9±0.5    \\
                    & UPNet (Ours)         & 82.3±0.7          & \textbf{91.3±0.4} & \textbf{94.5±0.2} & \textbf{97.1±0.1}  & \textbf{98.6±0.1}  & \textbf{2.2±0.1} & \textbf{0.9±0.0} & 0.4±0.1    \\ \bottomrule
    \end{tabular}}\label{Tab: comp}
\end{table*}

\begin{figure*}[!ht]
    \centering
    \subfloat[Sudbury]{\includegraphics[width=0.5\linewidth]{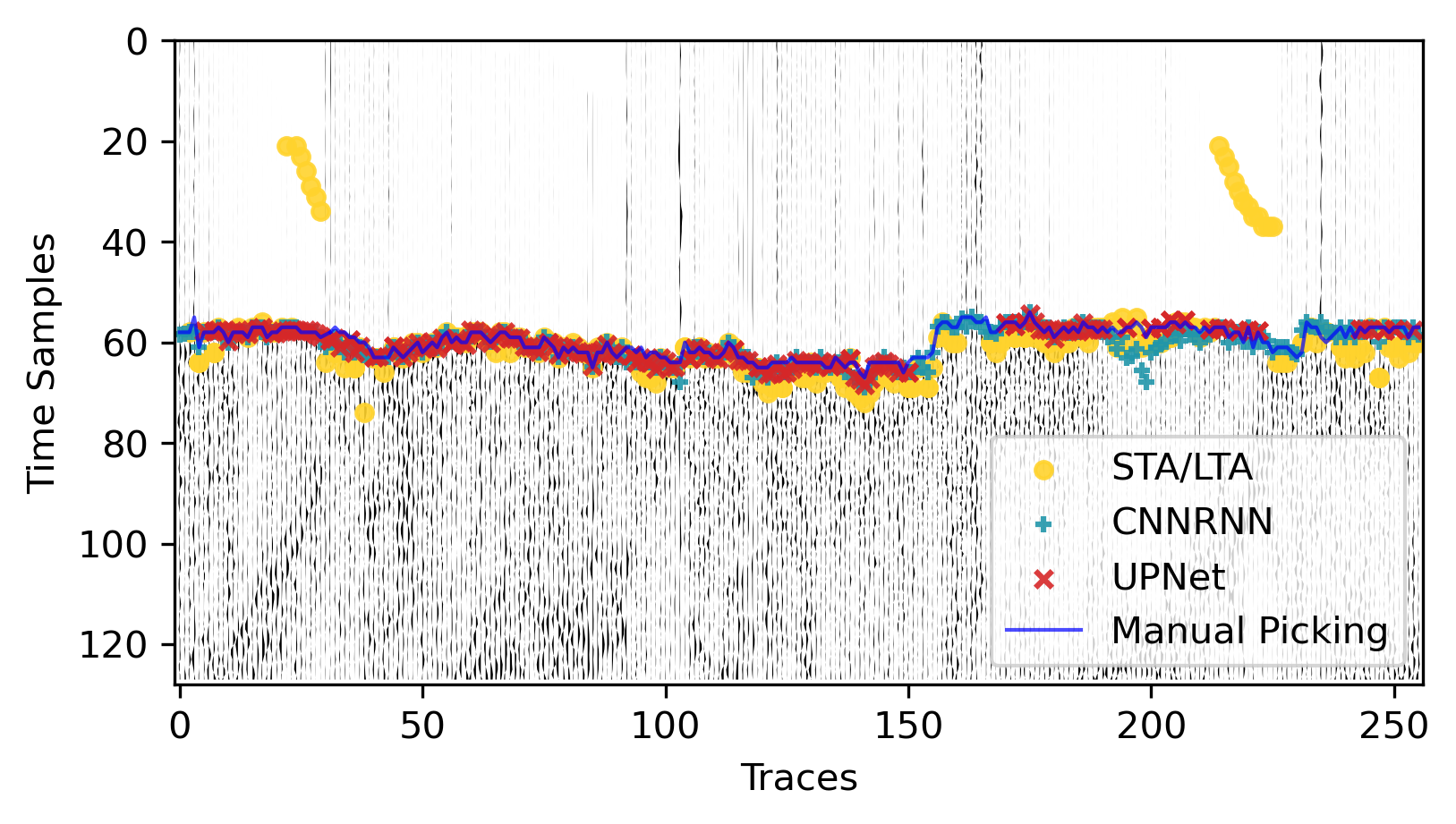}\label{fig: fold1}}\subfloat[Brunswick]{\includegraphics[width=0.5\linewidth]{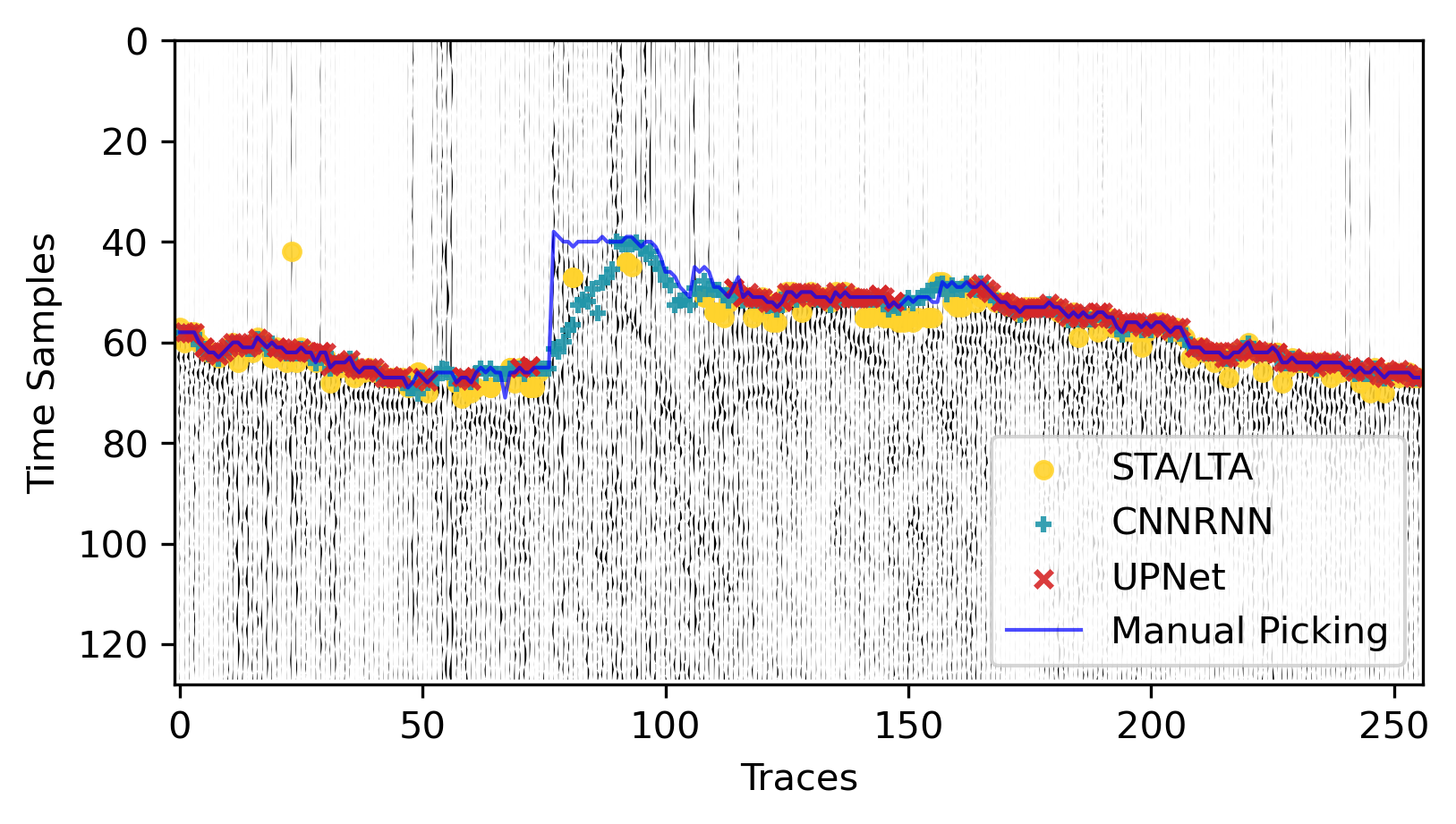}\label{fig: fold2}}\\
    \subfloat[Halfmile]{\includegraphics[width=0.5\linewidth]{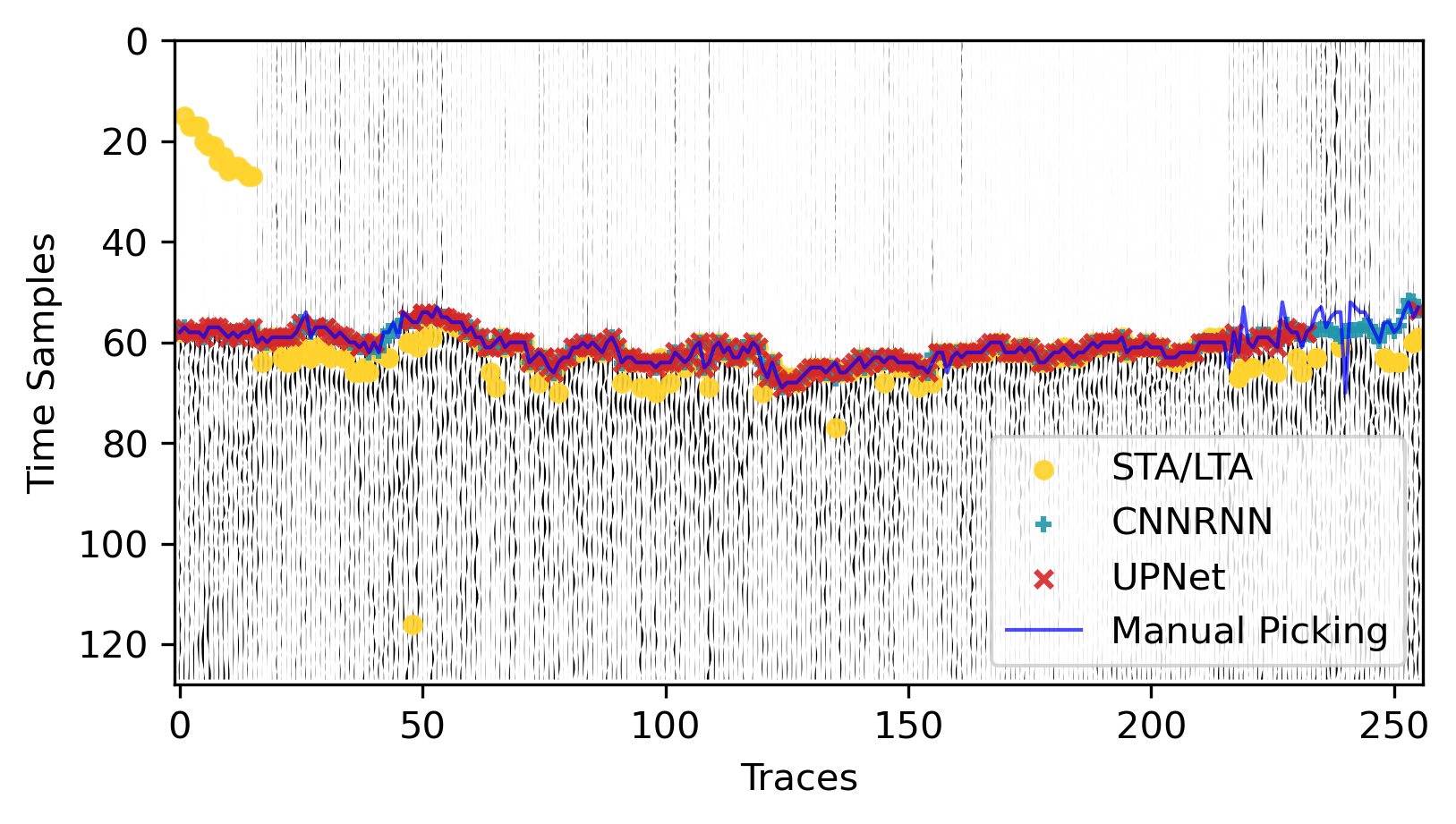}\label{fig: fold3}}\subfloat[Lalor]{\includegraphics[width=0.5\linewidth]{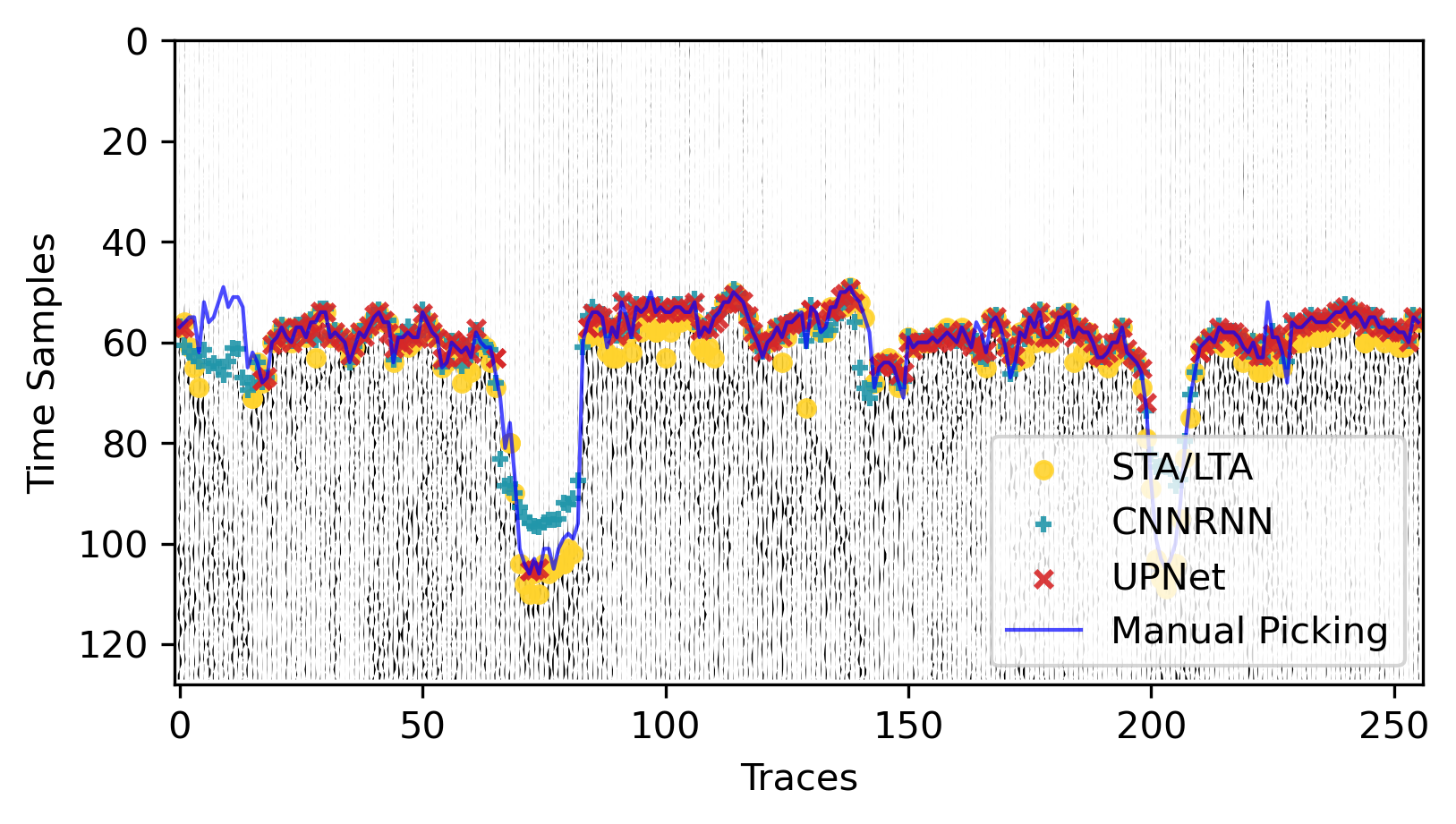}\label{fig: fold4}}
    \caption{Four classic picking cases of the STA/LTA, CNNRNN, UPNet, and manual picking on the surveys of Sudbury, Brunswick, Halfmile, and Lalor, respectively.}
    \label{fig: CompVisual}
\end{figure*}

\subsection{Comparative Experiments}
To evaluate the performance of UPNet, we compare four state-of-the-art (SOTA) methods for automatically picking methods quantitatively. We also visualize the picking results of a classic picking method, the current end-to-end SOTA method, and UPNet.
There are a few studies on the open-source dataset used in our study: the benchmark method\cite{st2024deep}, the self-trained network\cite{ozawa2024automated}, and the multi-stage segmentation picking network (MSSPN)\cite{wang2024msspn}.
Since the training method of benchmark\cite{st2024deep} and the self-trained network\cite{ozawa2024automated} are the same as ours, we compare UPNet with them.
UPNet is an end-to-end method. Thus, we also compare an end-to-end SOTA method called CNNRNN\cite{yuan2020robust}. Moreover, we also compared the traditional method STA/LTA to verify the generalization of automatic methods. 

Next, we detail the training processes of each comparative model. First, the length of the long window and the short window in the STA/LTA method are selected among $\{50, 40, 30, 20, 10\}$ and $\{5, 4, 3, 2\}$, respectively, as suggested by \cite{akram2016review}. We select the combinations of each fold corresponding to the best test MAE as the optimal parameters of STA/LTA.  
Second, due to the complexity of implementing the benchmark method and the self-trained network, we directly reference the experimental results in the original papers to prevent accuracy errors resulting from the issues we encountered during reproduction. Finally, we adopt the parameters used in the original paper to train CNNRNN—specifically, BS=4 for CNN, BS=10 for RNN, and Lr=1e-3 for both CNN and RNN. Moreover, the ground truth of the segmentation is involved in the benchmark method, the self-trained network, and CNNRNN. They all label the pixels before FBs as 0 and others as 1, similar to UPNet. 

For each fold, the benchmark method, the self-trained network, CNNRNN, and our UPNet are trained ten times with various seeds using the training hyperparameters mentioned above. Additionally, UPNet takes the sampling ten times in the inference process. 
Tab. \ref{Tab: comp} showcases the test results of these five models, where each value in the table indicates the mean value ± the standard deviation. Concretely, the APR of UPNet can be controlled by $t_{\text{var}}(p)$, and we assume 80\% traces can be picked. Thus, we set $t_{\text{var}}(0.8)$ to remove the unstable picking of UPNet. 
Tab. \ref{Tab: comp} illustrates that UPNet outperforms another four automatic picking methods on Brunswick, Halfmile, and Lalor test sites. 
Notably, the self-trained method is extraordinary with two round training processes. The incorrect manual picking is removed between the two training processes. There are a few lousy manual picking in Sudbury. Multiple training processes used in the self-trained network can correct these error. 
Thus, the self-trained network performs well in the HR@5-9px, RMSE, MAE of Sudbury. However, the precision of the self-training network is insufficient, so both UPNet and CNNRNN exceed it on the HR@1px and HR@3px.
UPNet performs better than the self-trained on accuracy (MAE) and stability (RMSE) for another three folds.
Then, UPNet can outperform STA/LTA, the benchmark, and CNNRNN a lot in the test of four-folds, specifically in fold 3. In the Brunswick test, the HR@1px of UPNet is 99\%, indicating its reliability in picking surveys with relatively high SNR in real-world scenarios.
Furthermore, comparing end-to-end methods (CNNRNN and UPNet) with the methods of the segmentation plus threshold post process (the benchmark and the self-trained network), we conclude that the end-to-end method is more suitable for the FB picking task. We analyze that the blurred boundary of the FB signal in the segmentation map causes the low HR@1px. Moreover, unlike CNNRNN, UPNet can filter unstable pickings based on computed uncertainty. Specifically, the lowest RMSE verifies that UPNet is the most robust method for picking FB. 
Finally, in comparing STA/LTA and deep learning-based methods, we observe that the accuracy and stability of STA/LTA can not meet the real need because of the many manual parameters and the various noises of the field. Therefore, the deep learning-based automatic picking method is better for the FB picking task. Through the above analysis, we have verified the superiority of UPNet from a quantitative perspective. Next, we visualize four classic picking cases of each test site.

Next, we visualize four classic picking cases on each test site, as shown in Fig. \ref{fig: CompVisual}. 
UPNet achieves relatively robust picking in these four examples, even in the Lalor dataset with low SNR. Concretely, Fig. \ref{fig: CompVisual}(d) shows that FB has continuous jumps, but UPNet achieves good recognition ability. Even in the places with large jumps (traces 60-80), UPNet picks a little robust FB. This verifies that UPNet has the picking characteristics of preferring to miss rather than to overuse.
Furthermore, UPNet can also be used to evaluate manual picking. For instance, at traces 230-250 in Fig. \ref{fig: CompVisual}(c), manual picking is terrible, but the picking of UPNet is still robust.
Moreover, UPNet can avoid over-picking. Specifically, in Fig. \ref{fig: CompVisual}(b), the noise at the position of trace 75-100 is extreme, and accurate FB cannot be easily picked manually. It is reasonable not to pick these traces in such a scenario. UPNet does this, but for CNNRNN, there is no measurement of the picking error. Finally, picking the STA/LTA method is acceptable in the areas where the FB signal is separated from the background noise. However, it is a local method. It cannot pick according to the global FB trend when the noise changes drastically. Therefore, the characteristic of local picking leads to easy jumping in picking, such as the jumping of yellow circles in Fig. \ref{fig: CompVisual}(a)-(c). Based on the analysis of the above visual results, we conclude that UPNet is more robust and suitable for practical applications than other methods.

\subsection{Uncertainty Representation in First Break Picking}
\begin{figure}[!ht]
    \centering
    \includegraphics[width=3.4in]{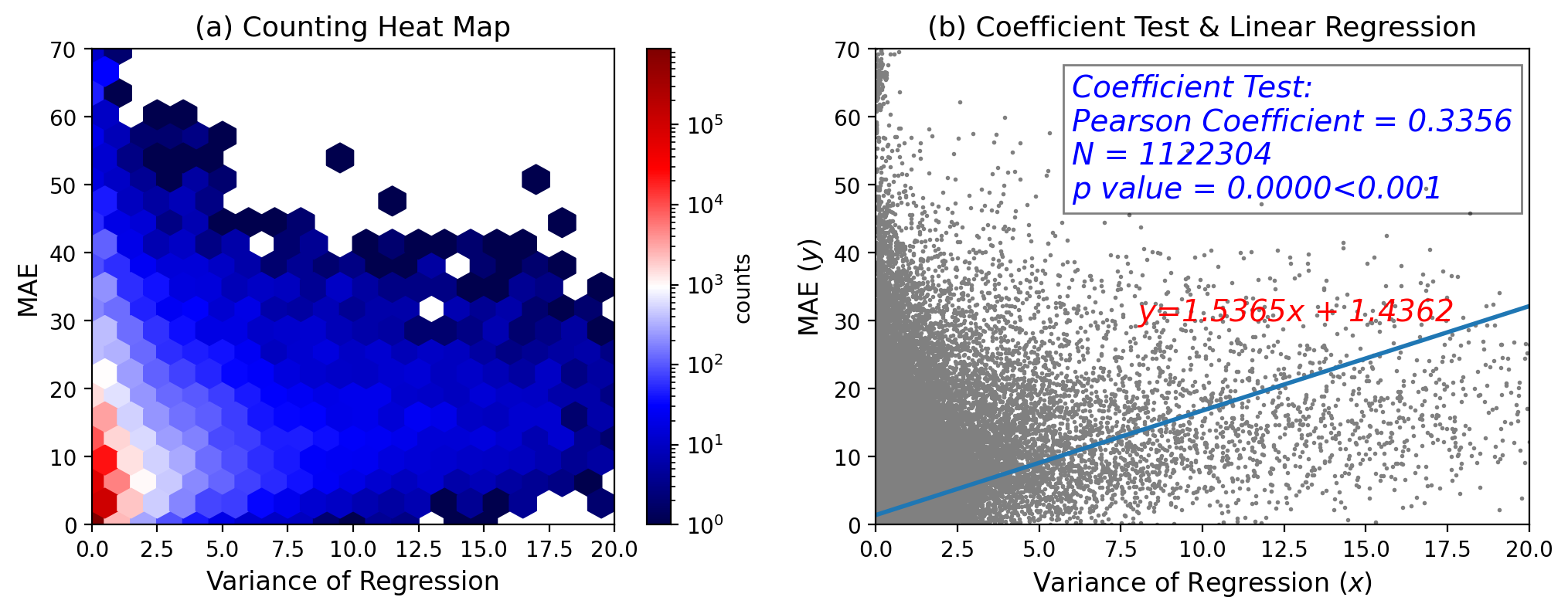}
    \caption{(a) Count the 2-dimensional points (variances of UPNet regression, MAE) and visualize them on a heat map. (b) Conduct Pearson coefficient test and linear regression.}
    \label{fig: uncer-global}
\end{figure}

After comparing the advantages of our method and other segmentation-based methods in evaluating metrics, we further explore why introducing uncertainty measures can improve the accuracy and stability of the first break picking. 
From the quantitative point of view, we calculate the sample distribution of UPNet output and analyze the correlation between the regression variances and corresponding MAEs. Specifically, we take a trained model on Fold 4 as an example. We take the variance of the sample distribution of UPNet regression as the horizontal coordinate and the MAE between the mean value of the sample distribution and the manual FB as the longitudinal coordinate. 
Fig. \ref{fig: uncer-global}(a) illustrates how these points are distributed on a plane and shows that a larger variance results in a bigger MAE. Further, we conduct linear regression based on these points to understand the quantifiable relationship between variance and MAE. Fig. \ref{fig: uncer-global}(b) indicates that the linear regression function is $y = 1.5365 x + 1.4362$, meaning that for every one increase in variance, MAE increases by 1.5365. Moreover, we also run the Pearson correlation test, and his null hypothesis is that the Pearson correlation coefficient $\rho$ is equal to 0, and his alternative hypothesis is that $\rho \neq 0$. In this showcase, the total number of points is 1122304, and the Pearson correlation coefficient $\rho=0.3356$. The hypothesis test corresponds to a p-value of 0.0000, rounded to 4 decimal places. Thus, we can reject the null hypothesis at a significance level of $\alpha=0.001$, meaning we consider it relevant. 

\begin{figure}[!ht]
    \centering
    \includegraphics[width=3.3in]{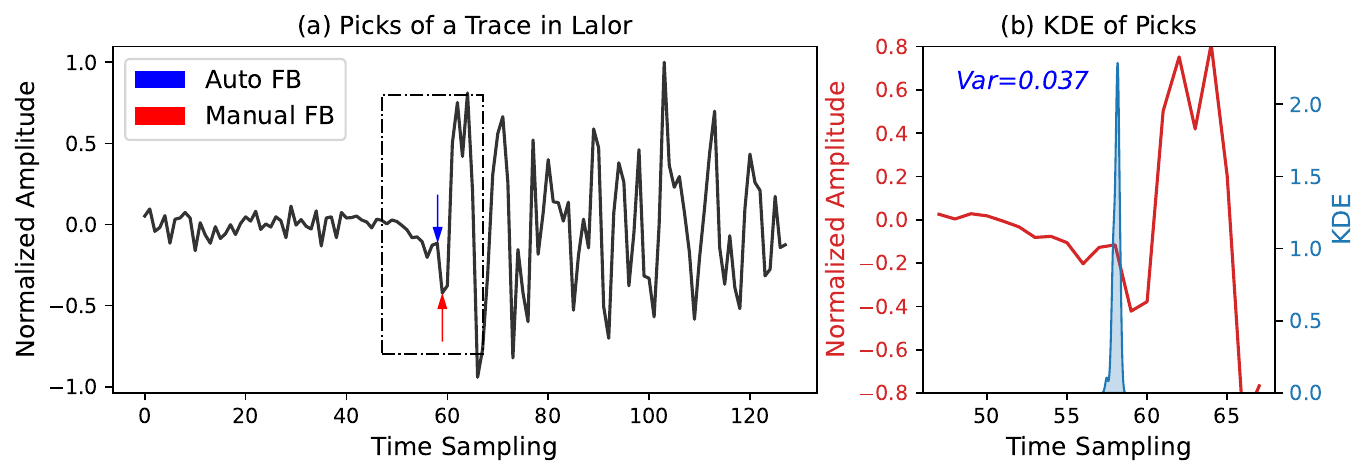}
    \caption{A showcase of the UPNet picking with a large variance.}
    \label{fig: uncer-bad}
\end{figure}

\begin{figure}[!ht]
    \centering
    \includegraphics[width=3.3in]{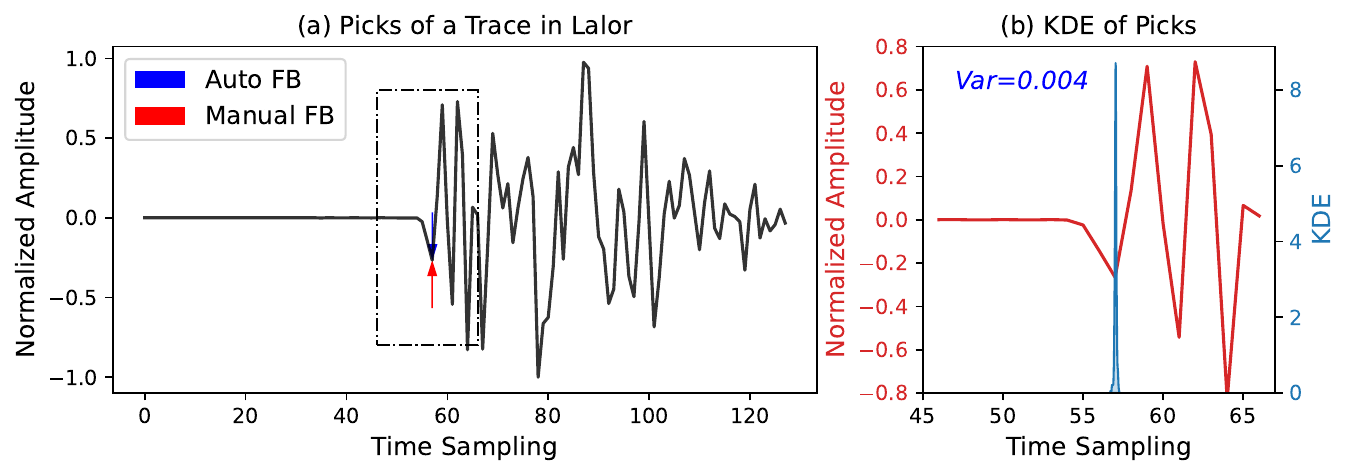}
    \caption{A showcase of the UPNet picking with a low variance.}
    \label{fig: uncer-good}
\end{figure}

In order to further verify the relationship between the estimated variance and the picking reliability, we visualize two specific picking cases on traces of Lalor as shown in Fig. \ref{fig: uncer-bad} and \ref{fig: uncer-good}. On the one hand, when the variance of the sampled distribution is large, the automatic picking of UPNet on this trace is suspect. Fig. \ref{fig: uncer-bad}(b) shows the kernel density estimation (KDE) of the sampled distribution of this trace, while Fig. \ref{fig: uncer-bad}(a) illustrates that the significant variance (0.037) leads to poor picking (indicated by the blue arrow). On the other hand, when the variance is slight, the picking of UPNet on this trace is reliable. Fig. \ref{fig: uncer-good} shows that picking with low variance in the sampled distribution is accurate.

\subsection{Robustness Test}
\begin{figure}[!ht]
    \centering
    \includegraphics[width=3.3in]{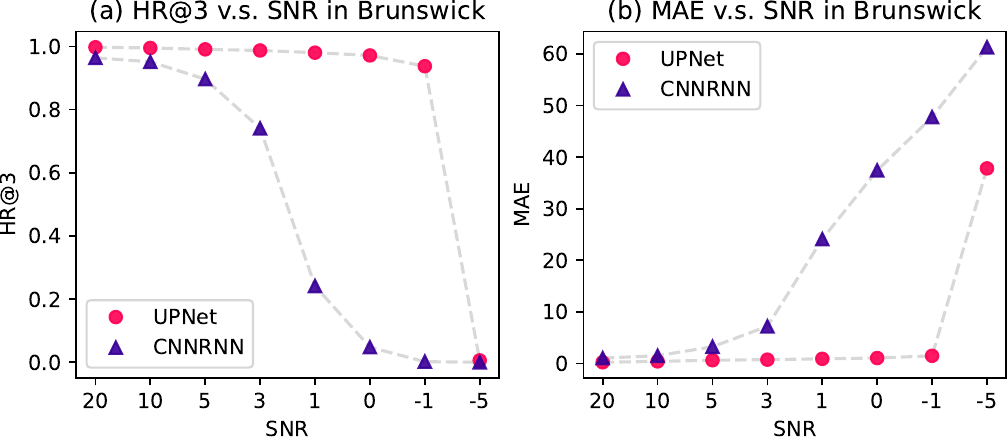}
    \caption{Anti-noise ability test of UPNet and CNNRNN on Brunswick.}
    \label{fig: robust}
\end{figure}

To further test the robustness of the UPNet, we add Gaussian noise with different SNRs to the test sites and evaluated both the trained UPNet and CNNRNN on the gather with various SNRs. We assume that the four test datasets used have relatively high SNRs, meaning no noise is present. Thus, we can inject Gaussian noise to generate the polluted gather with the expected constant SNR, which is defined by:
\begin{equation}
    \text{SNR} = 10 \times \log_{10}{\frac{\sigma_s^2}{\sigma_n^2}},
    \label{SNRDefine}
\end{equation}
where $\sigma_s^2$ and $\sigma_n^2$ represent the signal variances and the Gaussian noise, respectively. Subsequently, the variance of the added Gaussian noise can be computed using:
\begin{equation}
    \sigma_n^2 = \sigma_s^2 / 10^{\text{SNR}^*/10},
    \label{SNRSigmaN}
\end{equation}
where $\text{SNR}^*$ is the constant SNR and $\sigma_s^2$ can be estimated from each single trace signal directly.
We generate eight Gaussian noise data sets with 20, 10, 5, 3, 1, -1, and -5 SNRs, respectively. We select the best model in the training of ten random seeds for UPNet and CNNRNN in each fold experiment, totaling eight models. Then, we test these eight models on corresponding test datasets with various SNRs. For UPNet, we use the variance threshold $t_{var}(0.8)$ defined in Eq. \ref{UDM-thre} to filter the unstable pickings. Fig. \ref{fig: robust} visualizes the robustness test results on Brunswick. Specifically, UPNet outperforms CNNRNN under various SNRs, exhibiting higher HR@3px and lower MAE. In Fig. \ref{fig: robust}, we observe a phenomenon: CNNRNN shows a noticeable decrease in MAE at SNR=3, while for UPNet, it occurs at SNR=-1. This phenomenon also indicates that UPNet has more vital anti-noise ability than CNNRNN. 

\section{Conclusion}
In this paper, we propose a novel deep learning-based end-to-end FB picking framework that incorporates uncertainty quantification, thereby enhancing the accuracy and stability of the automatic picking process. After analyzing various experiments, three conclusions can be inferred. (1) UPNet can estimate the uncertainty of the picked FB directly and eliminate the unstable picks to enhance both accuracy and stability. (2) The test performance of UPNet in field surveys achieves SOTA, and the picking characteristics of UPNet also align with the principle of erring on the side of excess in practical applications. (3) Compared with other end-to-end deep learning methods such as CNNRNN, UPNet has more substantial anti-noise capabilities, further verifying its feasibility for application in field surveys.

\section*{Acknowledgment}
The author would like to thank Mr. Pierre-Luc St-Charles from Applied Machine Learning Research Team Mila, Québec AI Institute for providing the open datasets. 
\ifCLASSOPTIONcaptionsoff
  \newpage
\fi

\bibliographystyle{IEEEtran}
\bibliography{reference}


\end{document}